\documentclass[runningheads]{llncs}
\usepackage{times}
\usepackage{graphicx}
\usepackage{latexsym}
\usepackage{helvet}
\usepackage{courier}
\usepackage{amsmath}
\usepackage{amssymb}
\usepackage{algorithm}
\usepackage{alltt}
\usepackage{verbatim}
\usepackage{url}
\usepackage{xspace}
\usepackage[defblank]{paralist}

\usepackage{pgfplots}
\pgfplotsset{compat=1.16}
\usepackage{tikz}
\usepackage[caption=false]{subfig}

\usepackage{algorithmicx}
\usepackage{algpseudocode}

\newcommand{\ignore}[1]{}

\newcommand{\skipit}[1]{{ #1}}

\newcommand{\DL}{{ DL}}

\newcommand{\pl}{\partial_{||}}

\newcommand{\ARROW}{\hookrightarrow}

\newcommand{\supp}{\sigma}   
\newcommand{\non}{{\thicksim}}

\renewcommand{\imath}{i}

\newcounter{clause}
\def\theclause{$c$\arabic{clause}}

\newenvironment{clause}{\begin{tabbing}
xxx\=xxx\=xxx\=\+\kill}%
{\end{tabbing}}

\newenvironment{Clause}{\refstepcounter{clause}%
\begin{tabbing}
cxxxx\=xxx\=xxx\=\kill
\theclause\>\+}%
{\end{tabbing}}

\ignore{
\newtheorem{theorem}{Theorem}
\newtheorem{lemma}[theorem]{Lemma}
\newtheorem{definition}[theorem]{Definition}
\newtheorem{propn}[theorem]{Proposition}
\newtheorem{proposition}[theorem]{Proposition}
\newtheorem{corollary}[theorem]{Corollary}
\newtheorem{example}[theorem]{Example}
}

\newcommand{\mi}[1]{\mathit{#1}}

\ignore{
\newcounter{clause}
\def\theclause{$c$\arabic{clause}}

{\end{tabbing}}

{\end{tabbing}}

\newtheorem{theorem}{Theorem}
\newtheorem{lemma}[theorem]{Lemma}

\newtheorem{proposition}[theorem]{Proposition}
\newtheorem{corollary}[theorem]{Corollary}
\newtheorem{example}[theorem]{Example}
\newenvironment{proof}[1][Proof]{\begin{trivlist}\item[\hskip \labelsep {\bfseries #1}]}{\end{trivlist}}
}
\newtheorem{thm}{Alternative Theorem}

\begin{document}

\title{Approximating Defeasible Logics to Improve Scalability}

\author{Michael J. Maher} 
\institute{
Reasoning Research Institute \\
Canberra, Australia  \\
E-mail: michael.maher@reasoning.org.au
}

\maketitle

\begin{abstract}
Defeasible rules are used in providing computable representations of legal documents and, more recently,
have been suggested as a basis for explainable AI.
Such applications draw attention to the scalability of implementations.
The defeasible logic $DL(\pl)$
was introduced as a more scalable alternative to
$DL(\partial)$, which is better known.
In this paper we consider the use of (implementations of) $DL(\pl)$
as a computational aid to computing conclusions in $DL(\partial)$ and other defeasible logics, 
rather than as an alternative to $DL(\partial)$.
We identify conditions under which $DL(\pl)$ can be substituted for $DL(\partial)$
with no change to the conclusions drawn,
and conditions under which  $DL(\pl)$ can be used to draw some valid conclusions,
leaving the remainder to be drawn by $DL(\partial)$.

\keywords{defeasible logics \and approximation \and computation \and query optimization}

\end{abstract}

\section{Introduction}

Defeasible logics are systems for reasoning with defeasible rules, in a paraconsistent manner.
Defeasible rules have long been studied as a basis for formalizing
legal systems, documents, and reasoning \cite{Prakken,regs,contracts,Grosof04}.
More recently, defeasible rules have been proposed as a basis for providing
explanations for otherwise opaque AI systems \cite{VBP21,CT16,StrassWD19}.  
Such applications suggest the need for defeasible logic systems that can handle
defeasible rules on a large scale.

Despite \emph{propositional} defeasible logics having linear complexity  \cite{linear,TOCL10},
the implementation of conventional \emph{first-order} defeasible logics faces scalability issues \cite{ECAI2012,sdl}.
This led to the design of the defeasible logic $DL(\pl)$ \cite{sdl} as a more scalable alternative to
$DL(\partial)$, which is better known \cite{Nute87,BCN,Billington93,Nute_book,TOCL01}.
$DL(\pl)$ has particular advantages over $DL(\partial)$ \cite{sdl,sdl2},
which gives it greater potential for scalable implementation.
On the other hand, other defeasible logics have better intuitions and justifications for the reasoning they support \cite{Nute_book,ECAI00,TOCL10}.
In this paper we consider the use of (implementations of) $DL(\pl)$
as a \emph{computational aid} to computing with $DL(\partial)$  and other defeasible logics, 
rather than as an alternative to $DL(\partial)$.

We identify cases where $DL(\pl)$ can be directly substituted for another logic,
because they have the same consequences.
We also identify cases where the conclusions of $DL(\pl)$ are a subset of the conclusions of another defeasible logic -- $DL(\pl)$ is an under-approximation of the logic --
so that 
an implementation of $DL(\pl)$ can be used as a preprocessor for an implementation of that logic.
The cases where $DL(\pl)$ is an over-approximation of the logic are also of interest,
and we have some results there too.
The hope is that such results lead to the scalable implementation or approximation of
conventional defeasible logics, bypassing the bottlenecks referred to earlier.

The remainder of the paper is structured as follows.
Section~\ref{sec:defeasible_logic} introduces
concepts from defeasible reasoning,
and defines the defeasible logic $\DL(\pl)$ from \cite{sdl}.
Section~\ref{sec:decisive} identifies cases where a defeasible theory is decisive,
that is, every literal can either be defeasibly inferred ($D \vdash +d\: q$) or
provably not defeasible inferred ($D \vdash -d\: q$).
This information is used in later sections.
Section \ref{sec:equal} identifies cases where inference in $DL(\pl)$ is equivalent to inference
in other defeasible logics.  
Thus an implementation of $DL(\pl)$ can be substituted for an implementation of the other logic.
Section \ref{sec:contain} addresses cases where inference in $DL(\pl)$ is an approximation
of inferences in other logics, and thus an implementation of $DL(\pl)$ can be used to simplify
the processing needed by implementation of the other logic.
A brief discussion concludes the paper.

\section{Defeasible Logics}  \label{sec:defeasible_logic}
In this section we briefly introduce the basics of defeasible logics and establish notation.
For a fuller introduction to a variety of defeasible logics, we refer the reader to \cite{TOCL10}.

\subsection{Defeasible Theories}

A defeasible theory $D$ is a triple $(F,R,>)$ where $F$ is a finite set of facts (literals), 
$R$ a finite set of labelled rules,
and $>$ a superiority relation (a binary acyclic relation) on $R$ (expressed on the labels),
specifying when one rule overrides another, given that both are applicable.

A rule $r$ consists (a) of its antecedent (or body) $A(r)$ which is a finite set of literals, (b) an arrow, and, (c) its
consequent (or head) $C(r)$ which is a literal. Rules also have distinct \emph{labels} which are used to refer to
the rule in the superiority relation. There are three types of rules: strict rules, defeasible rules and
defeaters represented by their respective arrows: $\rightarrow$, $\Rightarrow$ and $\leadsto$. 
Strict rules are rules in the classical sense: whenever the premises are indisputable (e.g., facts) then so is the conclusion. 
Defeasible rules are rules that can be defeated by contrary evidence. 
Defeaters are rules that cannot be used to draw any conclusions; their only use is to provide contrary evidence that may prevent some conclusions.
We use $\ARROW$ to range over the different kinds of arrows used in a defeasible theory.

Rules may be written in a first-order syntax, but
each rule represents the set of its variable-free instances, and each variable-free atom is considered a proposition.
Thus, the defeasible theories are essentially propositional,
and there is no loss of generality in stating and proving results in propositional terms.
A literal is either a proposition or its negation.
Negation is represented by the symbol $\neg$,
and we define the complement operation $\non$ as follows:
if $q$ is a proposition $p$, then $\non q = \neg p$;
if $q$ has the form $\neg p$ then $\non q = p$.
A defeasible theory is \emph{hierarchical} (or  \emph{acyclic} or \emph{stratified}) if
there is a mapping $m$ which maps propositions to the non-negative integers such that, 
for every rule,
the head is mapped to a greater value than any proposition in the body, whether negated or not.
That is, there is no recursion in the rules of the defeasible theory,
not even through a literal's complement.

A defeasible theory is \emph{semi-hierarchical}  if
there is a mapping $m$ which maps literals to the non-negative integers such that, 
for every rule,
the head is mapped to a greater value than any body literal.
A defeasible theory is hierarchical iff it is semi-hierarchical via mapping $m$ and,
for every literal $q$, $m(q) = m(\non q)$.
Clearly a hierarchical defeasible theory is also semi-hierarchical.
A defeasible theory is \emph{strict semi-hierarchical} iff 
the set of strict rules is semi-hierarchical.

Given a rule $q_1, \ldots, q_n \ARROW p$, we say that the literal $p$ directly depends on each literal $q_i$.
Dependency is the transitive closure of direct dependency.
The notions of (semi-)hierarchy can also be formulated in terms of dependency.

Given a set $R$ of rules, we denote the set of all strict rules in $R$ by
$R_{s}$, and the set of strict and defeasible rules in $R$ by $R_{sd}$. $R[q]$ denotes the set of rules in $R$ with consequent $q$.

\begin{example}    \label{ex:tweety}
To demonstrate defeasible theories,
we consider the familiar Tweety problem and its representation as a defeasible theory.
The defeasible theory $D$ consists of the rules and facts
\[
\begin{array}{rrcl}
r_{1}: & \mi{bird}(X) & \Rightarrow & \phantom{\neg} \mi{fly}(X) \\
r_{2}: & \mi{penguin}(X) & \Rightarrow & \neg \mi{fly}(X) \\
r_{3}: & \mi{penguin}(X) & \rightarrow & \phantom{\neg} \mi{bird}(X) \\
r_{4}: & \mi{injured}(X)   & \leadsto     &  \neg \mi{fly}(X) \\
f      : & \mi{penguin}(\mi{tweety}) &  &  \\
g      : & \mi{bird}(\mi{freddie}) &  &  \\
h     : & \mi{injured}(\mi{freddie}) &  &  \\
\end{array}
\]
and a priority relation $r_{2} > r_{1}$.

Here $r_1, r_2, r_3, r_4, f$ are labels and
$r_3$ is (a reference to) a strict rule, while $r_1$ and $r_2$ are defeasible rules,
$r_4$ is a defeater,
and $f, g, h$ are facts.
Thus $F = \{f,g,h\}$, $R_s = \{ r_3 \}$ and $R_{sd} = R = \{r_1, r_2, r_3 \}$
and $>$ consists of the single tuple $(r_2, r_1)$.
The rules express that birds usually fly ($r_1$),
penguins usually don’t fly ($r_2$),
that all penguins are birds ($r_3$),
and that an injured animal may not be able to fly ($r_4$).
In addition, the priority of $r_{2}$ over $r_{1}$ expresses that when something is both a bird and a penguin (that is, when both rules can fire) it usually cannot fly
(that is, only $r_{2}$ may fire, it overrules $r_{1}$).
Finally, we are given the facts that $\mi{tweety}$ is a penguin,
and $\mi{freddie}$ is an injured bird.

This defeasible theory is hierarchical.
One function that demonstrates this
maps $\mi{injured}$ and $\mi{penguin}$ propositions to 0,
$\mi{bird}$ propositions to 1, and $\mi{fly}$ propositions to 2.
Among the literals, $\mi{fly(tweety)}$ depends on $\mi{penguin(tweety)}$
(via $\mi{bird(tweety)}$), while $\neg \mi{fly(tweety)}$ also depends on $\mi{penguin(tweety)}$.
\end{example}

\subsection{Defeasible Inference}
 
A defeasible logic is characterized by its inference rules.
When a literal is inferred, it is tagged by the inference rule that was used.
A \emph{conclusion} takes the form $+d \: q$ or $-d \: q$, where $q$ is a literal and $d$ is the tag indicating which inference rule was used.
Given a defeasible theory $D$,
$+d \: q$ expresses that $q$ can be proved via inference rule $d$ from $D$,
while $-d \: q$ expresses that it can be established within the logic that $q$ cannot be proved from $D$.
In this paper, we focus on $+d$ conclusions.
We write $D \vdash +d \: q$ to express that $+d \: q$ can be inferred from $D$
via a sequence of inferences (i.e. a proof).

Defeasible logics may involve several inference rules,
but usually have a principal inference rule that is intended to represent defeasible inference.
Such logics are denoted by $\DL( d )$,
where $d$ is the tag of the principal inference rule in the logic.
There are a variety of defeasible logics, several of which are defined and studied in \cite{TOCL10}.
$\DL(\partial)$ and $\DL(\partial^*)$ are ambiguity blocking, while
$\DL(\delta)$ and $\DL(\delta^*)$ are ambiguity propagating.
$\DL(\partial)$ and $\DL(\delta)$ employ team defeat, while
$\DL(\partial^*)$  and $\DL(\delta^*)$ use individual defeat.
For definition of these logics and discussion of these classifications, see \cite{TOCL10}.\footnote{
The defeasible inference rules we will refer to are denoted by
$\delta^*, \delta, \partial^*$, and $\partial$, 
while auxiliary (or support) inference rules are
$\supp_{\delta^*}, \supp_{\delta}, \supp_{\partial^*}$, and $\supp_{\partial}$,
following the notation of \cite{GM17} for support inference rules.
}
We present below the definition of the logic $\DL(\pl)$ from \cite{sdl},
which is ambiguity propagating and uses team defeat.

$\DL(\pl)$ involves three tags: 
$\Delta$, which expresses conventional monotonic inference;
$\lambda$, an auxiliary tag;
and $\pl$, which is the main notion of defeasible proof in this logic.
The inference rules are presented below, phrased as conditions on proofs.
Here,
$D$ is a defeasible theory  $(F,R,>)$;
$q$ is a variable-free literal;
$P$ denotes a proof, that is, a sequence of conclusions, 
each derived from previous conclusions by a single application of an inference rule;
$P[1..i]$ denotes the first $i$ elements of $P$;
and $P(i)$ denotes the $i^{th}$ element of $P$.
\begin{tabbing}
90123456\=7890\=1234\=5678\=9012\=3456\=\kill

$+\Delta$:  If  $P(\imath+1) = +\Delta q$  then either \\
\hspace{0.2in}  (1)  $q \in F$;  or \\
\hspace{0.2in}  (2)  $\exists r \in R_{s}[q] \  \forall a \in A(r),
+\Delta a \in P[1..\imath]$.
\end{tabbing}
\smallskip

This inference rule concerns reasoning about definitive information,
involving only strict rules and facts.
It is identical to the rule for monotonic inference in $\DL(\partial)$.

For a defeasible theory $D$,
we define $P_{\Delta}$ to be 
the set of consequences in the largest proof satisfying the proof condition $+\Delta$,
and call this the $\Delta$-\emph{closure}.
It contains all $+\Delta$ consequences of $D$.

Once $P_{\Delta}$ is computed, we can apply the $+\lambda$ inference rule.
$+\lambda q$ is intended to mean that $q$ is potentially defeasibly provable in $D$.
The $+\lambda$ inference rule is as follows.

\begin{tabbing}
$+\lambda$: \= If $P(\imath + 1) = +\lambda q$ then either \\
\> (1) \=$+\Delta q \in P_{\Delta}$ or \\
\> (2)	\>(2.1) $\exists r \in R_{sd}[q] ~ \forall \alpha \in A(r): +\lambda \alpha \in P(1..\imath)$ and \\
\> \>(2.2) $+\Delta \non q \notin P_{\Delta}$ 
\end{tabbing}
\smallskip

Using this inference rule, and given $P_{\Delta}$, we can compute the $\lambda$-closure $P_{\lambda}$.
which contains all $+\lambda$ consequences of $D$.

$+\pl q$ is intended to mean that $q$ is defeasibly provable in $D$.
Once $P_{\Delta}$ and $P_{\lambda}$ are computed, we can apply the $+\pl$ inference rule.

\begin{tabbing}
$+\pl$: \= If $P(\imath + 1) = +\pl q$ then either \\
\> (1) \=$+\Delta q  \in P_{\Delta}$ or \\
\> (2)	\>(2.1) $\exists r \in R_{sd}[q] ~ \forall \alpha \in A(r): +\pl \alpha \in P(1..\imath)$ and \\
\> \>(2.2) $+\Delta \non q \notin P_{\Delta}$ and \\
\> \>(2.3) \=$\forall s \in R[\non q]$ either \\
\> \> \>(2.3.1) $\exists \alpha \in A(s): +\lambda \alpha \notin P_{\lambda}$ or \\
\> \>\>(2.3.\=2) $\exists t \in R_{sd}[q]$ such that \\
\> \>\>\>$\forall \alpha \in A(t): +\pl \alpha \in P(1..\imath)$ and $t > s$
\end{tabbing}

The $\pl$-closure $P_{\pl}$ contains all $\pl$ consequences of $D$.

In the $+\pl$ inference rule,
(1) ensures that any monotonic consequence is also a defeasible consequence.
(2) allows the application of a rule (2.1) with head $q$, provided that
monotonic inference cannot prove $\non q$ (2.2)
and every competing rule either provably fails to apply (2.3.1)
or is overridden by an applicable rule for $q$ (2.3.2).
The $-\pl$ inference rule is the strong negation \cite{flexf} of the $+\partial$ inference rule.

Inference rules like $\pl$ employ the notion of ``team defeat'',
where it doesn't matter which rule overrides an opposing rule, as long as all opposing rules are overridden.
This is expressed in (2.3.2).
We can also have a version of $\pl$ with ``individual defeat'',
where all opposing rules must be overridden by the same rule,
which we denote by $\pl^*$.
The inference rule for $+\pl^*$ replaces (2.3.2) in $\pl$ by the simple condition $r > s$.

\begin{example}    \label{ex:tweety3}
We now apply these
inference rules to the Tweety defeasible theory in Example \ref{ex:tweety}. 

The $+\Delta$ inference rule infers
$+\Delta\: \mi{penguin}(\mi{tweety})$,
$+\Delta\: \mi{bird}(\mi{freddie})$, and \linebreak
$+\Delta\: \mi{injured}(\mi{freddie})$ 
from the facts, and
$+\Delta\: \mi{bird}(\mi{tweety})$
using $r_3$.

Using the $+\lambda$ inference rule
we infer all the literals inferred by the $+\Delta$ inference rule as $+\lambda$ conclusions.
In addition, the rule infers
$+\lambda\: \mi{fly}(\mi{tweety})$,
$+\lambda\: \mi{fly}(\mi{freddie})$, and
$+\lambda\: \neg\mi{fly}(\mi{tweety})$.

Using the $+\pl$ inference rule,
again all the $+\Delta$ conclusions are inferred as $+\pl$ conclusions.
The only other conclusion that can be drawn with this rule is
$+\pl\: \neg\mi{fly}(\mi{tweety})$.
The potential inference of $\mi{fly}(\mi{tweety})$ is overruled by $r_2$ inferring $+\pl\: \neg\mi{fly}(\mi{tweety})$.
On the other hand, a potential inference of $\mi{fly}(\mi{freddie})$ is not obtained
because $r_2$ cannot overrule $r_4$.
\end{example}

For proof tags $d_1$ and $d_2$, 
we say $d_2$ has \emph{greater inference power} than $d_1$ if,
for any defeasible theory $D$ and literal $q$,
$D \vdash +d_1 \, q$ then $D \vdash +d_2 \, q$.
That is, $d_2$ can infer any literal that $d_1$ can infer.
We will use the notation $+d_1 \subseteq +d_2$.
We will also use this notation when $D$ is restricted in certain ways, that is,
$d_2$ can infer any literal that $d_1$ can infer on all defeasible theories $D$ that satisfy the restriction.
The restriction should be clear from the context.
\cite{sdl} has a comprehensive statement of (unrestricted) relative inference power
among the inference rules we will address.

\section{Decisiveness}  \label{sec:decisive}

It turns out that notions of decisiveness are important to express conditions
under which $\DL(\pl)$ can (partially) compute the conclusions of a conventional defeasible logic.
Given a tag $d$, a defeasible theory is \emph{$d$-decisive} iff, for every literal $q$,
either $D \vdash +d\, q$ or $D \vdash -d\, q$.
In general, defeasible theories are not $d$-decisive,
by the nature of defeasible rules \cite{TOCL10}.

Theorem 2.1 of \cite{TPLP06} says (in different terminology)
\begin{theorem}[\cite{TPLP06}]     \label{thm:hier1}
Let $D$ be a finite propositional defeasible theory.

\noindent
If $D$ is hierarchical then $D$ is $\Delta$-decisive and $\partial$-decisive.
\end{theorem}

We will extend this result to other inference rules.
For $\Delta$ and $\lambda$ we can ensure decisiveness with a weaker condition.
\begin{proposition}    \label{prop:semihier}
Let $D = (F, R, >)$ be a finite propositional defeasible theory.

\noindent
If $D$ is semi-hierarchical then $D$ is $\Delta$-decisive and $\lambda$-decisive.
\end{proposition}
\skipit{
\begin{proof}
Let $m$ be the layer function.
The proof is by induction on the layer of the literal.
Suppose $D$ is $\Delta$-decisive on all literals $q$ with $m(q) < k$.
Consider a literal $p$ with $m(p) = k$.

If $p \in F$ then $D \vdash +\Delta p$.
Otherwise, if $R$ contains a strict rule with head $p$
and, for every literal $q$ in the body, $D \vdash +\Delta q$ then,
by $+\Delta.2$, $D \vdash +\Delta p$.
Otherwise, every strict rule with head $p$ contains a literal $q$ in the body with $D \vdash -\Delta q$,
by the hierarchy on $D$ and the induction hypothesis.
Hence, by the $-\Delta$ inference rule, $D \vdash -\Delta p$.
Thus $p$ is $\Delta$-decided by $D$.
By induction, all literals are $\Delta$-decided by $D$, that is, $D$ is $\Delta$-decisive.

The proof for $\lambda$-decisive is similar, except for handling clause (2.2).
If $D \vdash +\Delta p$, then $D \vdash +\lambda p$, by clause (1).
Otherwise, if $D \vdash +\Delta \non p$ then $D \vdash -\lambda p$ by $-\lambda.2.2$.
Otherwise, we must have $D \not\vdash +\Delta \non p$. 
If $R$ contains a strict or defeasible rule with head $p$
and, for every literal $q$ in the body, $D \vdash +\lambda q$ then,
by $+\lambda.2$, $D \vdash +\lambda p$.
If not, then every strict or defeasible rule with head $p$ contains 
a literal $q$ in the body with $D \vdash -\lambda q$,
by the hierarchy on $D$ and the induction hypothesis.
Thus $p$ is $\lambda$-decided by $D$.
By induction, all literals are $\lambda$-decided by $D$, that is, $D$ is $\lambda$-decisive.
\end{proof}
}

We can slightly strengthen this result by looking at only the relevant parts of $D$.
Indeed, if $( \emptyset, R_s, \emptyset)$ is semi-hierarchical then $D$ is $\Delta$-decisive.
Similarly, if $( \emptyset, R_{sd}, \emptyset)$ is semi-hierarchical then $D$ is $\lambda$-decisive.

We now extend Theorem \ref{thm:hier1} to many other inference rules.
\begin{theorem}    \label{thm:hier2}
Let $D$ be a finite propositional defeasible theory.

For each
$d \in \{ \delta^*, \delta, \partial^*, \partial, \pl^*, \pl, \supp_{\delta^*}, \supp_{\delta}, \supp_{\partial^*}, \supp_{\partial} \}$,
if $D$ is hierarchical then $D$ is $d$-decisive.
\end{theorem}
\skipit{
\begin{proof}
The proof is by induction on the layers of the hierarchy.
Suppose $D$ is $d$-decisive on all literals $q$ with $m(q) < k$, for all such $d$.
Consider a literal $p$ with $m(p) = k$.

If $D \vdash +\Delta p$, then $D \vdash +d p$, by clause (1).
Otherwise, we must have $D \vdash -\Delta p$,
since $D$ is $\Delta$-decisive, so clause (1) of the inference rule $-d$ is satisfied.
For $d \in \{\partial^*, \partial, \delta^*, \delta \}$,
if $D \not\vdash -\Delta \non p$, then $D \vdash +\Delta \non p$,
since $D$ is $\Delta$-decisive.
By clause (2.2) of $-d$, we must have $D \vdash -d p$.
For $d \in \{\pl^*, \pl \}$, if $D \not\vdash +\Delta \non p$ does not hold, then $D \vdash +\Delta \non p$ holds so,
by clause (2.2) of $-d$, we must have $D \vdash -d p$.
(This discussion is irrelevant to inference rules $\supp_d$ because they omit the corresponding clause.)

Now, for any $d$, consider the rules for $p$ and $\non p$.
Because $D$ is hierarchical, and by the induction hypothesis,
all literals in the body of these rules are $d$-decided and $d'$-decided, where $d'$ is the tag referenced in
clause (2.3.1) of the defeasible inference rules (clause (2.2.1) of the support inference rules).
Suppose clause (2) of $+d$ is satisfied by some rule $r$.
Then $D \vdash +d p$.
Otherwise, clause (2) of $-d$ is satisfied, since the two conditions are negations of each other,
given that $D$ is $d$-decisive on the literals in the bodies of rules of $p$
and is $d'$-decisive on the literals in the bodies of rules of $\non p$.
Thus $D \vdash -d p$.
Hence $p$ is $d$-decided by $D$.
By induction, all literals are $d$-decided by $D$, that is, $D$ is $d$-decisive.
\end{proof}
}

Note that, although $D$ is $d$-decisive for many different $d$'s,
that does not mean that the different $d$'s agree on their conclusions.
We can demonstrate this using standard examples.
\begin{example}  \label{ex:dec}
Consider $D$ to be 

\[
\begin{array}{lrcl}
r: &           & \Rightarrow & \phantom{\neg} q \\
s: &           & \Rightarrow & \phantom{\neg} q \\
t: &           & \Rightarrow & \neg q \\
u: &           & \Rightarrow & \neg q \\
\end{array}
\]
with superiority relation $r > t$ and $s > u$.

Then $D \vdash +\partial q$, but $D \vdash -\partial^* q$,
and similarly for $\delta$ compared to $\delta^*$.

As another example, suppose $D$ is
\[
\begin{array}{lrcl}
r: &    p      & \Rightarrow & \phantom{\neg} q \\
s: &            & \Rightarrow & \neg q \\
t: &           & \Rightarrow & \phantom{\neg} p \\
u: &           & \Rightarrow & \neg p \\
\end{array}
\]
with empty superiority relation.
Then $D \vdash +\partial \neg q$ but $D \vdash -\delta \neg q$,
and similarly for $\partial^*$ compared  to $\delta^*$.
\end{example}

We will need to address some properties of individual literals 
that cause $D$ to be non-hierarchical or not semi-hierarchical.
We say a literal $q$ is \emph{conflicted} if there is a rule for $q$ and a rule for $\non q$.
When there are no conflicted literals, the universal quantifier in clause (2.3) of
inference rules $+\pl$ and $\pl^*$ is vacuously true.
It follows that, in this case, $\DL(\pl)$ and $\DL(\pl^*)$ agree.
A \emph{looping literal} is a literal $q$ such that $q$ or $\non q$ depends on either $q$ or $\non q$.
That is, if we treat $q$ and $\non q$ as a single item then there is a self-dependency.
If a literal $q$ depends on $q$ then we call it a \emph{self loop}.
A \emph{strict loop} is a self loop that depends on itself via only strict rules.

There is a close relationship between the existence of hierarchies and loops.
\begin{proposition}   \label{prop:hierloop}
Let $D$ be a finite propositional defeasible theory.

\begin{itemize}
\item
$D$ is hierarchical iff $D$ has no looping literal
\item
$D$ is semi-hierarchical iff $D$ has no self loop
\item
$D$ is strict semi-hierarchical iff $D$ has no strict loop
\end{itemize}
\end{proposition}
\skipit{
\begin{proof}
1.  
If $D$ is hierarchical then there is a layer mapping $m$ such that
if $p$ depends on $q$ then $m(p) > m(q)$ and $m(p) = m(\non p)$, for all $p$ and $q$.
If $D$ has a looping literal $p$ then $p$ or $\non p$ depends on $p$ or $\non p$.
Since $m(p) = m(\non p)$, we have  $m(p) > m(p)$ -- a contradiction.
Thus if $D$ is hierarchical then $D$ has no looping literal.
Conversely, if $D$ has no looping literal then we can topologically sort the predicates,
and then construct a mapping $m$ from the resulting sequence of predicates.

2.  
Suppose $D$ is  semi-hierarchical.
Then there is a layer mapping $m$ such that
if $p$ depends on $q$ then $m(p) > m(q)$, for all $p$ and $q$.
If $D$ has a  looping literal $p$ then $p$ depends on $p$,
and so $m(p) > m(p)$ -- a contradiction.
Thus if $D$ is semi-hierarchical then $D$ has no self loop.
Conversely, if $D$ has no self loop then we can topologically sort the literals,
and then construct a mapping $m$ from the resulting sequence of literals.

3.  
Let $D = (F, R, >)$ and $D' = (\emptyset, R_s, \emptyset)$.
$D$ is strict semi-hierarchical iff $D'$ is semi-hierarchical.
Furthermore, $D'$ is semi-hierarchical iff $D'$ has no self loop, by part 2.
Since $D'$ consists only of $R_s$,
$D'$ has no self loop iff $D$ has no strict loop.
Thus, $D$ is strict semi-hierarchical iff $D$ has no strict loop.
\end{proof}
}

Combining results earlier in this section with Proposition \ref{prop:hierloop}, we have
\begin{corollary}   \label{cor:loopdec}
Let $D$ be a finite propositional defeasible theory, and let \\
$d \in 
\{ \delta^*, \delta, \partial^*, \partial, \pl^*, \pl, \supp_{\delta^*}, \supp_{\delta}, \supp_{\partial^*}, \supp_{\partial} \}$.
\begin{itemize}
\item
If $D$ has no looping literal, then $D$ is $d$-decisive
\item
If $D$ has no self loops, then $D$ is $\lambda$-decisive
\item
If $D$ has no strict loops, then $D$ is $\Delta$-decisive
\end{itemize}
\end{corollary}
\skipit{
\begin{proof}
1.
By Proposition \ref{prop:hierloop}, if $D$ has no looping literal then $D$ is hierarchical.
Then, by Theorem \ref{thm:hier2}, $D$ is $d$-decisive.

2.
By Proposition \ref{prop:hierloop}, if $D$ has no self loops then $D$ is semi-hierarchical.
Now, by Proposition \ref{prop:semihier}, $D$ is $\lambda$-decisive.

3.
Let $D = (F, R, >)$ and $D' = (F, R_s, \emptyset)$.
By Proposition \ref{prop:hierloop}, if $D$ has no strict loops then $D$ is strict semi-hierarchical.
Hence $D'$ is semi-hierarchical.
Now, by Proposition \ref{prop:semihier}, $D'$ is $\Delta$-decisive
and, hence, so is $D$ (since they have the same facts and strict rules).
\end{proof}
}

As a particular case, if $R_s = \emptyset$ then $D$ is $\Delta$-decisive.
The following definition is used later.
A defeasible theory $D = (F, R, >)$ is \emph{fact-deficient} if 
$F = \emptyset$ and every rule in $R_s$ has a non-empty body.
For such defeasible theories there are no definitive (i.e. positive, strict) consequences,
that is, no $q$ such that $D \vdash +\Delta q$.
These theories are not necessarily $\Delta$-decisive, because they may contain
strict loops.

Obviously, the results in this section are based  on the syntactic structure of $R$.
We can refine this kind of analysis by considering more details of $D$ and the logic.
For example, if $p \in F$ then all rules for $p$ and non-strict rules for $\non p$ could be deleted from $R$
producing an equivalent (in terms of consequences) defeasible theory with fewer dependencies.
As another example, if $p \notin F$ and $p$ is not the head of any rule, then
any rule containing $p$ in the body can be deleted from $R$.

Finally, on decisiveness, while it is clear that $\Delta$-decisiveness does not imply $d$-decisiveness,
the issue of whether $d$-decisiveness implies $\Delta$-decisiveness is less clear.
The following example shows it does not, in general.

\begin{example}   \label{ex:partdec}
We show that it is possible for a defeasible theory $D$ to be $\partial$-decisive,
but not $\Delta$-decisive.
Consider $D$ to be
\[
\begin{array}{lrcl}
r: &     q      & \rightarrow & q \\
s: &           & \Rightarrow &  q \\
\end{array}
\]
with empty superiority relation.

Then $D$ is not  $\Delta$-decisive because neither $+\Delta q$ nor $-\Delta q$ is a consequence of $D$.
On the other hand, $-\Delta \non q$ is a consequence.
Consequently, so is $-\partial \non q$ and $+\partial q$.
Thus $D$ is $\partial$-decisive.

The same argument applies to any of the logics we address in this paper.
\end{example}

When a literal is $\partial$-decided but $\Delta$-undecided
there are further consequences that are entailed.
\begin{lemma}  \label{lemma:undec}
Let $D$ be a defeasible theory.
If $D$ is $\partial$-decisive and a literal $p$ is $\Delta$-undecided,
then the following are consequences of $D$:

\begin{itemize}
\item $-\Delta \non p$
\item $+\partial p$
\item $-\partial \non p$
\item $+\lambda p$
\end{itemize}
\end{lemma}
\skipit{
\begin{proof}
For the literal $p$, neither $+\Delta p$ nor $+\Delta p$ is a consequence of $D$.
Then clause 1 of the $-\partial$ inference rule is not satisfied for $p$.
Since $D$ is $\partial$-decisive, $+\partial p$ must be a consequence.
Similarly, clause 1 of the $+\partial$ inference rule is not satisfied for $p$,
so clause 2 must be satisfied and, in particular, clause 2.2.
Thus $-\Delta \non p$ is a consequence of $D$.

Now, clause 1 of the $+\partial$ inference rule is not satisfied for $\non p$,
because we know $-\Delta \non p$ is a consequence.
But also clause 2.2 of the $+\partial$ inference rule is not satisfied for $\non p$,
because we know $-\Delta p$ is not a consequence.
Thus, the $+\partial$ inference rule cannot be applied for $\non p$.
Since $D$ is $\partial$-decisive, $-\partial \non p$ must be a consequence.

Finally, because $+\partial \subseteq +\lambda$ \cite{sdl}, we also have $+\lambda p$.
\end{proof}
}
The same argument applies for other inference rules, because they share the same structure.

\section{$\DL(\pl)$ as a Substitute for other logics}   \label{sec:equal}

If we think of $\DL(\pl)$ as an approximation to $\DL(\partial)$,
rather than a logic in its own right,
it becomes interesting to identify cases where the two logics agree.
In such cases, $\DL(\pl)$ is a scalable substitute for  $\DL(\partial)$.
It was already shown in \cite{sdl} (Theorem 11)
that the $+\pl$ inference rule makes different inferences, in general,
than other defeasible logics,
with neither stronger than the other.
We now look to identify a case where $\DL(\pl)$ makes the same inferences as other logics.

We first address a simple case where $\DL(\pl)$  and other logics agree.
When there are no conflicted literals, 
defeasible logics reduce to a variant of definite clauses.
More technically,
when clause (2.1) is satisfied,
the universal quantifier in clause (2.3) of
defeasible inference rules such as $+\pl$ is vacuously true,
as is clause (2.2) of support inference rules.
Furthermore, whether rules are strict or defeasible makes no difference when there is no conflict,
and defeaters have no effect.
As a result, all these inference rules 
agree on the consequences of these defeasible theories.

\begin{lemma}   \label{lemma:equal}
Let $D$ be a finite propositional defeasible theory with no conflicted literals.  

\noindent
Let 
$d \in \{ \delta^*, \delta, \partial^*, \partial, \pl^*, \pl, \supp_{\delta^*}, \supp_{\delta}, \supp_{\partial^*}, \supp_{\partial}, \lambda \}$.

\noindent
Then all inference rules $+d$ behave identically (as do all inference rules $-d$).
Specifically, for all literals $q$ and all tags $d$ and $d'$:
\begin{itemize}
\item
$D \vdash +d q$ iff $D \vdash +d' q$
\item
$D \vdash -d q$ iff $D \vdash -d' q$
\end{itemize}
\end{lemma}
\skipit{
\begin{proof}
For any of the positive  inference rules,
in the absence of conflicts,  if clause (2.1) is satisfied then
clause (2.3) of the defeasible inference rules 
(and clauses (2.2) of the support inference rules) are satisfied.
This is because the universal quantifier in (2.3)  is vacuously satisfied.
Furthermore, 
clause (2.2) of the defeasible inference rules is satisfied because,
in the absence of conflicts, there can be no rule for $\non q$
and hence there is an immediate inference of $-\Delta \non q$.

Thus all positive inference rules, whether defeasible or support,
reduce to (1) and (2.1).
Consequently, all the inference rules have the same structure, effectively,
and so all prove the same literals.

The argument is dual, but essentially the same, for the negative inference rules.
The existential quantifier in (2.3) cannot be satisfied,
and (2.2) cannot hold because there is no rule for $\non q$.
\end{proof}
}

Indeed, any reasonable defeasible logic can be expected to agree with these logics
on conflict-free defeasible theories.

We now extend Lemma~\ref{lemma:equal} to a somewhat more general class of theories.
\begin{theorem}     \label{thm:equiv}
Let $D$ be a finite propositional defeasible theory.
Suppose no conflicted literal depends on a conflicted literal,
and no conflicted literal depends on a looping literal.
Then
\begin{itemize}
\item
$\DL(\delta)$, $\DL(\partial)$ and $\DL(\pl)$ agree on the conclusions 
from $D$.
\item
$\DL(\delta^*)$, $\DL(\partial^*)$ and $\DL(\pl^*)$ agree on the conclusions 
from $D$.
\end{itemize}
\end{theorem}
\skipit{
\begin{proof}
First, consider unconflicted literals that do not depend on a conflicted literal.
By Lemma~\ref{lemma:equal}, 
all the inference rules agree  on all such literals.

Now, consider a conflicted literal $q$ in $D$ and the set of rules $R[q]$ and $R[\non q]$.
Since no conflicted literal depends on a looping literal,
the literals in the bodies of the rules $R[q] \cup R[\non q]$ are defined by a hierarchical set of rules and,
because no conflicted literal depends on a conflicted literal, none of the literals $q$ and $\non q$ depend on are conflicted.
Thus, by Lemma~\ref{lemma:equal}, all of these literals and, in particular,
all of the literals in the bodies of the rules $R[q] \cup R[\non q]$, 
are accorded the same status by all the inference rules.
Thus, apart from (2.3.1), all clauses are evaluated the same in $+\pl$ and $+\partial$ (and also in $-\pl$ and $-\partial$) inference rules.

For any literal $\alpha$ as mentioned in (2.3.1), we have
$-\supp_{\delta} \alpha \in P_\delta$ iff $-\partial \alpha \in P_\partial$ iff $-\lambda \alpha \in P_\lambda$, by Lemma~\ref{lemma:equal}, 
and $-\lambda \alpha \in P_\lambda$ iff $+\lambda \alpha \notin P_\lambda$, by Theorem \ref{thm:hier2}.
Thus (2.3.1) will be evaluated the same in  $+\delta$ , $+\partial$, and $+\pl$  inference rules.
Similarly, 
(2.3.1) will be evaluated the same in $-\supp_{\delta} $,  $-\partial$, and $-\pl$  inference rules.
Thus $\DL(\delta)$, $\DL(\partial)$, and $\DL(\pl)$ agree on conclusions involving $q$ and $\non q$.

Finally, consider unconflicted literals that depend on a conflicted literal $q$.
Although $\delta$, $\partial$ and $\pl$ agree on $q$, $\lambda$ may differ.
However, 
inference for unconflicted literals only involves clauses (1) and (2.1) of the respective inference rules,
as noted in the proof of Lemma~\ref{lemma:equal}.
Consequently, the difference has no effect.  
Following the same argument as in the proof of Lemma~\ref{lemma:equal},
the three inference rules $+\delta$, $+\partial$ and $+\pl$ make the same inferences 
(as do $-\delta$, $-\partial$ and $-\pl$)
on the remaining unconflicted literals.

Thus $\DL(\delta)$, $\DL(\partial)$ and $\DL(\pl)$ agree on conclusions from $D$.

By essentially the same argument $\DL(\delta^*)$, $\DL(\partial^*)$ and $\DL(\pl^*)$ agree on conclusions from $D$.
\end{proof}
}

The Tweety defeasible theory (Example~\ref{ex:tweety}) satisfies the conditions of this theorem, so it can be directly computed by $\DL(\pl)$.
The first part of Example~\ref{ex:dec} shows that the two parts of this theorem
produce different conclusions, in general.

The limitations on the use of conflicted literals make this proposition of limited use.
Defeasible theories often have cascades of conflicted literals that are disallowed
by the conditions of this proposition.
Nevertheless, it might be possible to restructure such theories in the appropriate form:
a cascade of conflicted literals might be replaced by independent conflicts, that are then conjoined.
We will not pursue this thought further, here, but an example demonstrates the idea.

\begin{figure}[t]
\begin{minipage}{.45\textwidth}
\begin{center}
\[
\begin{array}{lrcl}
&   q             & \Rightarrow & r \\
\\
\\
&   s             & \Rightarrow & \neg q \\
&  p, v         & \Rightarrow & \phantom{\neg} q \\
&  t              & \Rightarrow & \neg p \\
&  u              & \Rightarrow & \phantom{\neg} p \\
\end{array}
\]
(a)
\end{center}
\end{minipage}
\begin{minipage}{.45\textwidth}
\begin{center}
\[
\begin{array}{lrcl}
&   q', p          & \Rightarrow & \phantom{\neg} r \\
&   s              & \Rightarrow & \neg q' \\
&   v             & \Rightarrow & \phantom{\neg} q' \\
&   s             & \Rightarrow & \neg q \\
&  p, v         & \Rightarrow & \phantom{\neg} q \\
&  t               & \Rightarrow & \neg p \\
&  u              & \Rightarrow & \phantom{\neg} p \\
\end{array}
\]
(b)
\end{center}
\end{minipage}
\label{fig:restructure}
\caption{Restructuring a defeasible theory to satisfy the condition of Theorem~\ref{thm:equiv}.}
\end{figure}

\begin{example}    \label{ex:}  
Consider the defeasible theory
\ignore{
\[
\begin{array}{lrcl}
&   q             & \Rightarrow & r \\
&   s             & \Rightarrow & \neg q \\
&  p, v         & \Rightarrow & \phantom{\neg} q \\
&  t              & \Rightarrow & \neg p \\
&  u              & \Rightarrow & \phantom{\neg} p \\
\end{array}
\]
}
in Figure~\ref{fig:restructure}(a),
with additional rules for $s$, $t$, $u$, $v$ and their negations.

Then $q$ is a conflicted literal that depends on the conflicted literal $p$.
Consequently, Theorem~\ref{thm:equiv} cannot be used to conclude that the two logics agree on $r$.
However, this theory could be re-written as
in Figure~\ref{fig:restructure}(b).
\ignore{
\[
\begin{array}{lrcl}
&   q', p          & \Rightarrow & \phantom{\neg} r \\
&   s              & \Rightarrow & \neg q' \\
&   v             & \Rightarrow & \phantom{\neg} q' \\
&   s             & \Rightarrow & \neg q \\
&  p, v         & \Rightarrow & \phantom{\neg} q \\
&  t               & \Rightarrow & \neg p \\
&  u              & \Rightarrow & \phantom{\neg} p \\
\end{array}
\]
}
Now $q'$ does not depend on $p$, and the previous result applies.
We conclude that $r$ has the same status in the two logics.

This is a particularly simple situation, and it is not straightforward to extend it to more complicated cases
and to first-order syntax.
Furthermore, it can lead to an explosion of rules.
Nevertheless, it might be a viable approach when the rule-base is relatively small
and the fact-base is large.
\end{example}

An interesting corollary of the previous theorem is that it gives us a case where 
$\partial^*$ is weaker in inference strength than $\partial$ (which does not hold in general \cite{TOCL10}).

\begin{corollary}   
Let $D$ be a finite propositional defeasible theory.
If no conflicted literal depends on a conflicted literal,
and no conflicted literal depends on a looping literal,
then 

$D \vdash +\partial q$ if $D \vdash +\partial^* q$.
\end{corollary}
\skipit{
\begin{proof}
By Theorem 11 of \cite{sdl}, if $D \vdash +\pl^* q$ then $D \vdash +\pl q$.
Applying Theorem~\ref{thm:equiv} gives us the result.
\end{proof}
}

Theorem~\ref{thm:equiv} is far from a characterization of when two logics agree.
It is easy to find examples where the effect of a loop is masked.
For example
\[
\begin{array}{lrcl}
r: &  p, \mi{false}  & \Rightarrow & q \\
s: &  p                  & \Rightarrow & p \\
\end{array}
\]
where $\mi{false}$ is a literal with no other occurrence in $D$,
and
\[
\begin{array}{lrcl}
r: &    & \Rightarrow & p \\
s: &  p      & \Rightarrow & p \\
\end{array}
\]
Similarly, the effect of conflicted literals can be masked.
For example,
\[
\begin{array}{lrcl}
r: &  \mi{false}  & \Rightarrow & \neg p \\
s: &                  & \Rightarrow & \phantom{\neg} p \\
\end{array}
\]
and, when there is a fact $q \in F$,
\[
\begin{array}{lrcl}
r: &  & \Rightarrow & \neg q \\
s: &  & \Rightarrow & \phantom{\neg} q \\
\end{array}
\]
Thus $\DL(\partial)$ and $\DL(\pl)$ agree on many defeasible theories
that are not captured by Theorem~\ref{thm:equiv}.

Notwithstanding this point, we present two examples demonstrating the need for the conditions in Theorem~\ref{thm:equiv}.
The examples work for both parts of Theorem~\ref{thm:equiv}.
The first example shows that $\DL(\partial)$ and $\DL(\pl)$ can differ if a conflicted literal depends on a looping literal.
\begin{example}    \label{ex:confloop}  
Consider the defeasible theory
\[
\begin{array}{lrcl}
r: &           & \Rightarrow & \neg q \\
s: &  p      & \rightarrow & \phantom{\neg} q \\
t: &  \neg p      & \rightarrow & \neg p \\
\end{array}
\]

Here $q$ is a conflicted literal, depending on the looping literal $p$.
Then $P_\Delta = \emptyset$ and $P_\lambda = \{ +\lambda \neg q \}$.
Consequently $+\pl \neg q$ can be concluded.
On the other hand, we cannot conclude $+\partial \neg q$,
because we cannot conclude $-\partial p$.
This effect doesn't change if $s$ and $t$ are defeasible rules, instead of strict.
\end{example}

The second example shows that $\DL(\partial)$ and $\DL(\pl)$ can differ if a conflicted literal depends on a conflicted literal.

\begin{example}      \label{ex:partialpl}  
Consider the defeasible theory
\[
\begin{array}{lrcl}
r: &           & \Rightarrow & \phantom{\neg} q \\
s: &           & \Rightarrow & \neg q \\
t: &           & \Rightarrow & \phantom{\neg} p \\
u: &     q    & \Rightarrow & \neg p \\
\end{array}
\]
with no superiority relation.

Then we can infer $-\partial q$ and $+\lambda q$.
Consequently, we can infer $+\partial p$, but not $+\pl p$.
Hence, $\partial \not\subseteq \pl$.
This difference would still hold if there were intervening unconflicted literals linking $q$ and $\neg p$.
It comes about because the inference rules for $+\pl$ and $+\partial$ differ at (2.3.1):
$+\pl$ requires $+\lambda \alpha \notin P_\lambda$ while $+\partial$ requires $-\partial \alpha \in P(1..i)$.
\end{example}
\noindent
As a side note, this example also demonstrates that $\DL(\pl)$ is an ambiguity-propagating logic.

\section{$\DL(\pl)$ as a Preprocessor}   \label{sec:contain}

In situations where the relationship between $\pl$ and another tag $d$ is well-defined,
we may be able to exploit that relationship to compute the $d$-consequences of a theory
using $\pl$.
In particular, we address under- and over-approximations.
Let $D$ be a defeasible theory, $d$ be a proof tag, and let $C_d = \{ q~|~ D \vdash +d q \}$
be the $d$-closure.
A set of literals $S$ is an \emph{under-approximation} of $d$ if $S \subseteq C_d$,
and an \emph{over-approximation} of $d$ if $C_d \subseteq S$.
Thus $+\pl \subseteq +d$ expresses that the closure $C_{\pl}$ is an under-approximation of $d$.
In such situations, $\DL(\pl)$ can serve as a scalable pre-processor,
eliminating many literals from consideration before employing $\DL(d)$.

Any under-approximation can help us compute $C_d$.

\begin{proposition}
Let $D$ be a defeasible theory, $d$ be a proof tag, and let $C_d$ be the $d$-closure.

If $S$ is an under-approximation of $d$ then
the closure of $S$ under the application of the $+d$ inference rule (and auxiliary rules) is $C_d$.
\end{proposition}
\skipit{
\begin{proof}
Straightforward.
\end{proof}
}

That is,  an under-approximation $S$ to $d$ can be used to ``kick-start'' inference,
leaving the $+d$ inference rule to only infer the remainder of $C_d$. 
We now identify circumstances where $\pl$ can be used to compute an under-approximation
of some inference rules, beginning with $\partial$.

\begin{theorem}     \label{thm:contain1}
Let $D$ be a $\partial$-decisive defeasible theory.
Then $+\pl \subseteq +\partial$.
\end{theorem}
\skipit{
\begin{proof}
 Suppose, to obtain a contradiction,
that there is a literal $q$ such that $+\pl q$ is a consequence of $D$, but $+\partial q$ is not.
Choose $q$ to be such a literal with the shortest proof of  $+\pl q$.
Since $D$ is $\partial$-decisive, $-\partial q$ is a consequence.

We now show that the applicability of the $+\pl$ inference rule to infer $+\pl q$
and the applicability of the $-\partial$ inference rule to infer $-\partial q$ leads to a contradiction.

Because $-\partial q$ is a consequence of $D$, by clause $-\partial.1$, $-\Delta q$ is a consequence of $D$.
Thus $+\pl.1$ is not applicable for $q$ and, hence, $+\pl.2$ is applicable.
In particular, by $+\pl.2.2$, $+\Delta \non q$ is not a consequence and, hence,
$-\partial.2.2$ does not hold.

Also, by $+\pl.2.1$, there is a strict or defeasible rule $r$ where, for all $a \in A(r)$,
$+\pl a$ is a consequence.
Since $+\pl a$ must have a shorter proof than $+\pl q$, and given the choice of $q$ as the literal
with the shortest proof of those such that $+\pl q$ is a consequence  but $+\partial q$ is not,
we must have $+\partial a$, for all $a \in A(r)$.
Hence, $-\partial.2.1$ does not hold, by the coherence of $DL(\partial)$ \cite{TOCL10}.

By $+\pl.2.3$, for every rule $s$ for $\non q$, either
$+\lambda b$ is not a consequence, for some $b \in A(s)$
or there is a strict or defeasible rule $t$ for $q$ where $t > s$ and,
for each $c \in A(t)$, $+\pl c$ is a consequence of $D$.
In the first case, since $+\partial \subseteq +\lambda$ \cite{sdl}, $-\partial.2.3.1$ does not hold for such $s$.
In the second case, again by the choice of $q$ and because $+\pl c$ must have a shorter proof than $+\pl q$,
$+\partial c$ must be a consequence of $D$, for each $c \in A(t)$.
It follows that $-\partial.2.3$ also does not hold.

But the inapplicability of $-\partial.2.1$, $-\partial.2.2$, and $-\partial.2.3$
contradicts the applicability of the $-\partial$ inference rule to $q$.
Thus our original supposition is incorrect and
if $+\pl q$ is a consequence of $D$ then $+\partial q$ is also a consequence.

\end{proof}
}

A more straightforward proof, by induction on the length of proofs in $DL(\pl)$,
is available if we add the extra condition that $D$ is $\Delta$-decisive.

\begin{thm}     \label{thm:contain1.2}
Let $D$ be a finite propositional defeasible theory.
Suppose that $D$ is $\Delta$-decisive and $\partial$-decisive.
Then $+\pl \subseteq +\partial$.
\end{thm}
\skipit{
\begin{proof}
Before we begin the proof, we establish two facts used in the proof.
Let $\alpha$ be a literal.
First, if $+\Delta \non q \notin P_\Delta$ then, by the $\Delta$ decisiveness of $D$,
$-\Delta \non q \in P_\Delta$.
Second, by Theorem 11 of \cite{sdl}, $\partial \subset \lambda$
and hence if $+\lambda \alpha \notin P_\lambda$ then $+\partial \alpha \notin P_\partial$,
for any literal $\alpha$.
By the  $\partial$-decisiveness of $D$, 
if $+\lambda \alpha \notin P_\lambda$ then $-\partial \alpha \in P_\partial$.

Let $q$ be a literal.
The proof is by induction on the length of the proof of $+\pl q$ in $\DL(\pl)$.
If the length of the proof of $q$ is 2 then $q$ must be a fact, and then also $D \vdash +\partial q$.
Suppose $+\pl \subseteq +\partial$ for all literals $p$ such that a proof of $\pl p$ has length less than $k$.
Consider $q$, where $+\pl q$ has smallest proof of length $k$ and 
consider the application of the inference rule $+\pl$ to prove $+\pl q$.
We show that there is a proof in $\DL(\partial)$ of $+\partial q$.

If $+\pl.1$ applies, then so does $+\partial.1$.
Otherwise $+\pl.2$ applies.
The literals $\alpha$ in $+\pl.2.1$ must have a proof of length less than $k$
and hence, by the induction hypothesis, there is a proof of $+\partial \alpha$.
Thus  $+\partial.2.1$ can be satisfied.
$+\pl.2.2$ applies, so $+\Delta \non q \notin P_\Delta$ and hence, as noted above, $-\Delta \non q \in P_\Delta$.
Hence $+\partial.2.2$ can be satisfied.
$+\pl.2.3$ applies so, for each $s$ either $+\pl.2.3.1$ or $+\pl.2.3.2$ applies.
If $+\pl.2.3.1$ applies to $s$ using $\alpha$ then, by the second fact noted above,
$-\partial \alpha$ can be proved and hence $+\partial.2.3.1$ can be satisfied.
If $+\pl.2.3.2$ applies then $+\pl \alpha$ has proof of length less than $k$ so,
by the induction hypothesis, $+\partial \alpha$ can also be proved ,
and hence $+\partial.2.3.1$ can be satisfied.
Thus $+\partial q$ can be proved.

Thus, by induction, if $+\pl q$ can be proved from $D$, so can $+\partial q$.
That is, $+\pl \subseteq +\partial$.
\end{proof}
}

Applying Theorem  \ref{thm:hier1}, we have the following corollary to Theorem~\ref{thm:contain1},
which provides a syntactic counterpart of the theorem.
\begin{corollary}     \label{cor:contain1}
Let $D$ be a finite propositional defeasible theory.
If $D$ is hierarchical then $+\pl \subseteq +\partial$.
\end{corollary}
\skipit{
\begin{proof}
By Theorem \ref{thm:hier1}, $D$ is $\Delta$-decisive and $\partial$-decisive.
Applying the previous proposition, $+\pl \subseteq +\partial$.
\end{proof}
}

With essentially the same proof as Theorem~\ref{thm:contain1}, and its corollary, 
we have a similar result for inference using individual defeat.
\begin{proposition}     \label{prop:contain2}
Let $D$ be a finite propositional defeasible theory.

If $D$ is $\partial^*$-decisive, then $+\pl^* \subseteq +\partial^*$.

In particular, if $D$ is hierarchical then $+\pl^* \subseteq +\partial^*$.
\end{proposition}

Containment is strict for the three previous results.
That is, for example, there is a defeasible theory $D$ and literal $q$ such that
$D$ is $\partial$-decisive, and $+\partial q$ is a consequence,
but  $+\pl q$ is not.
This can be seen from a second look at the defeasible theory in Example \ref{ex:partialpl}.
\begin{example}
Consider again Example \ref{ex:partialpl}.
We have consequences $-\partial \neg q$, $-\partial q$, $-\partial \neg p$, and $+\partial p$ in $\DL(\partial)$.
Hence $D$ is $\partial$-decisive, as well as clearly $\Delta$-decisive.
But $+\pl p$ is not a consequence in $\DL(\pl)$.
Thus the containment is strict.
\end{example}

The $\partial$-decisiveness requirement in Theorem~\ref{thm:contain1}  is necessary.
\begin{example}   \label{ex:thm4p2}
We show that when $D$ is not $\partial$-decisive,
even if it is $\Delta$-decisive,
the conclusion of Theorem~\ref{thm:contain1} may not hold.
Consider $D$ to be
\[
\begin{array}{lrcl}
s: &           & \Rightarrow & \neg q \\
t: &    q      & \Rightarrow & \phantom{\neg} q \\
\end{array}
\]
with empty superiority relation.

Then $D$ is $\Delta$-decisive: we have $-\Delta q$ and $-\Delta \neg q$.
$D$ is not $\partial$-decisive because we do not have $+\partial \neg q$ nor $-\partial \neg q$.
Within $DL(\pl)$,
we have $+\lambda \neg q$, but not $+\lambda q$.
It follows that $+\pl \neg q$ is a consequence, since  2.1, 2.2, and 2.3.1 of the $+\pl$ inference rule are satisfied.
Thus $+\pl \not\subseteq +\partial$, for this defeasible theory.
\end{example}


We now turn to the ambiguity propagating logics.
The support inference rules $\supp_{\delta}$ and $\supp_{\delta^*}$ play a similar role,
in defining $\delta$ and $\delta^*$, as $\lambda$ plays in defining $\pl$ and $\pl^*$.
For the ambiguity propagating logics, a difficulty in obtaining similar results
to Theorem~\ref{thm:contain1} and Proposition~\ref{prop:contain2}
arises from the fact that $+\supp_d \subseteq +\lambda$ is not true, in general,
where $d$ is one of the defeasible tags.
When it holds, that condition enables us to establish that
if $+\lambda \alpha \notin P_\lambda$ (2.3.1 of the $+\pl$ inference rule) 
then $-\supp_d \alpha$ is provable (2.3.1 of the $+d$ inference rule).
Hence, we need to find sufficient conditions to ensure $+\supp_d \subseteq +\lambda$.
Fact-deficiency is such a condition.

\begin{lemma}     \label{lemma:supplambda}
Let $D$ be a fact-deficient finite propositional defeasible theory,
and let $d \in \{\delta^*, \delta, \partial^*, \partial \}$.
Then $+\supp_d \subseteq +\lambda$.
\end{lemma}
\skipit{
\begin{proof}
First note that, as a result of fact-deficiency, there are no conclusions of the form $+\Delta p$,
that is,  $+\Delta p \notin P_\Delta$, for every literal $p$.
Now, the proof is by induction on the length of proof of conclusions $+\supp_d  q$.
If $+\supp_d$.1 can be applied, so can $+\lambda$.1.
If $+\supp_d$.2.1 can be applied, so can $+\lambda$.2.1, using the induction hypothesis.
$+\lambda$.2.2 is always satisfied, since $+\Delta \non q  \notin P_\Delta$.
Thus, by induction, any proof of $+\supp_d q$ can be imitated to prove $+\lambda q$.
\end{proof}
}

Under an extra condition on $D$, $\lambda$ is the same as the other support inference rules.
\begin{corollary}     \label{cor:supplambda}
Let $D$ be a fact-deficient finite propositional defeasible theory
with empty superiority relation.
Let $d \in \{\delta^*, \delta, \partial^*, \partial \}$.
Then $+\supp_d = +\lambda$.
\end{corollary}
\skipit{
\begin{proof}
If the superiority relation of $D$ is empty then $\supp_d$.2.2 is satisfied.
Added to the proof in the previous lemma,
the induction shows that
$\supp_\delta$ and $+\lambda$ perform the same.
\end{proof}
}

The next result shows that $\pl$ ($\pl^*$) is an under-approximation of 
$\delta$ (respectively, $\delta^*$), under some conditions.
This also suggests that, under these conditions, the approximation of $\partial$ ($\partial^*$) 
in Theorem~\ref{thm:contain1} (respectively, Proposition~\ref{prop:contain2}) is weak,
because it is known from \cite{TOCL10} that
$\delta$ (respectively, $\delta^*$) under-approximates $\partial$ (respectively, $\partial^*$).

\begin{proposition}     \label{prop:contain3}
Let $D$ be a fact-deficient finite propositional defeasible theory.
\begin{itemize}
\item
If $D$ is $\supp_{\delta}$-decisive and $\Delta$-decisive,  then $+\pl \subseteq +\delta$.
\item
If $D$ is $\supp_{\delta}$-decisive and $\delta$-decisive,  then $+\pl \subseteq +\delta$.
\item
If $D$ is $\supp_{\delta^*}$-decisive and $\Delta$-decisive,  then $+\pl^* \subseteq +\delta^*$.
\item
If $D$ is $\supp_{\delta^*}$-decisive and $\delta^*$-decisive,  then $+\pl^* \subseteq +\delta^*$.
\end{itemize}
\end{proposition}
\skipit{
\begin{proof}
The proof of part 1 is by induction on the length of the proof of $+\pl q$ in $\DL(\pl)$,
as in Alternative Theorem~\ref{thm:contain1.2}.
Let $d \in \{\delta, \delta^* \}$.
$+\pl.1$ and $+\delta.1$ are identical.
By $\Delta$-decisiveness, if $+\pl.2.2$ holds then $+\delta.2.2$ holds.
Since $+\supp_\delta \subseteq +\lambda$, by Lemma~\ref{lemma:supplambda}, 
if $+\lambda \alpha \notin P_\lambda$ then $+\supp_\delta \alpha$ is not provable.
Since $D$ is $\supp_{\delta}$-decisive, 
$-\supp_\delta \alpha$ must be provable.
Thus, if $+\pl.2.3.1$ holds then so does $+d.2.3.1$.
Consequently, any inference made by $\pl$ can also be made by $\delta$.
Essentially the same argument handles part 3, dealing with $\pl^*$ and $\delta^*$.

The proof of parts 2 and 4 follows the proof of Theorem~\ref{thm:contain1}.
We suppose $+\pl q$ is a consequence and $+d q$ is not, for some $q$.
We use $d$-decisiveness to establish that $-d q$ is provable.
By Lemma~\ref{lemma:supplambda}, $+\supp_d \subseteq +\lambda$.
By $\supp_d$-decisiveness, $-\delta.2.3.1$ does not apply.
Hence we obtain a contradiction, as in the proof of Theorem~\ref{thm:contain1}.
\end{proof}
}
By Theorem~\ref{thm:hier2}, we have as a corollary that if $D$ is hierarchical and fact-deficient then
$+\pl \subseteq +\delta$ and $+\pl^* \subseteq +\delta^*$.

\begin{corollary}     \label{cor:hierAP}
Let $D$ be a hierarchical fact-deficient finite propositional defeasible theory.
Then $+\pl \subseteq +\delta$ and $+\pl^* \subseteq +\delta^*$.
\end{corollary}
\skipit{
\begin{proof}
Let $d \in \{\delta, \delta^* \}$.
By Theorem~\ref{thm:hier2}, $D$ is $d$-decisive and $\supp_d$-decisive.
Now, by Proposition~\ref{prop:contain3}, $+\pl \subseteq +\delta$ and $+\pl^* \subseteq +\delta^*$.
\end{proof}
}

Using Corollary~\ref{cor:supplambda},
we can identify cases where $\pl$ can substitute for $\delta$
(and $\pl^*$ for $\delta^*$).
\begin{proposition}     \label{prop:contain4}
Let $D$ be a fact-deficient, $\Delta$-decisive, finite propositional defeasible theory
with empty superiority relation.
\begin{itemize}
\item
If $D$ is $\supp_{\delta}$-decisive or $\lambda$-decisive, then $+\pl = +\delta$.
\item
If $D$ is $\supp_{\delta^*}$-decisive or $\lambda$-decisive, then $+\pl^* = +\delta^*$.
\end{itemize}
\end{proposition}
\skipit{
\begin{proof}
By Corollary~\ref{cor:supplambda}, $+\supp_d = +\lambda$.
Now, by $\lambda$- or $\supp_d$-decisiveness,
$+\lambda \alpha \notin P_\lambda$ iff $-\supp_d \alpha$ is provable.
The remainder of the proof follows that of Proposition~\ref{prop:contain3}.
\end{proof}
}

Since $+\lambda$ is the simpler inference rule, it seems likely that establishing decisiveness will be easier for $\lambda$. 

We also have:
\begin{proposition}     \label{prop:contain5}
Let $D$ be a finite propositional defeasible theory.
\begin{itemize}
\item
If $D$ is $\delta^*$-decisive, then $+\pl^* \subseteq +\supp_{\delta^*}$.
\item
If $D$ is $\delta$-decisive, then $+\pl \subseteq +\supp_{\delta}$.
\end{itemize}

\end{proposition}
\skipit{
\begin{proof}
By Theorem 11 of \cite{sdl}, $+\delta^* \subset +\lambda$
and hence if $+\lambda \alpha \notin P_\lambda$ then $+\delta^* \alpha \notin P_{\delta^*}$,
for any literal $\alpha$.
By the  $\delta^*$-decisiveness of $D$, 
if $+\lambda \alpha \notin P_\lambda$ then $-\delta^* \alpha \in P_{\delta^*}$.

Also note that $r > s$ implies $\neg(s > r)$, because $>$ is acyclic.

Using these facts, the proof by induction is similar to that of Alternative Theorem~\ref{thm:contain1.2},
based on the length of proofs of $+\pl^* q$.

For the second part, a similar argument shows that
if $+\lambda \alpha \notin P_\lambda$ then $-\delta \alpha \in P_{\delta}$.
In addition, note that if every rule for $\non q$ is inferior to a rule for $q$
in some set $S$ of rules for $q$ and $\non q$, 
then there is a rule for $q$ that is not inferior to any rule for $\non q$.
This because $>$ is acyclic.
Consequently, if $+\pl.2.3$ holds  then $+\supp_{\delta}.2.2$ holds.

Again, we can apply induction on the length of proofs of $+\pl q$ to obtain this result.
\end{proof}
}

Finally, we turn to over-approximations to $d$.
Given an over-approximation $S$, we can prune it so that it is closed under $d$ by
deleting elements of $S$ that cannot be inferred by $+d$ from $S$.
Repeated deletions will eventually lead to a $d$-closed set,
but this set is not necessarily the \emph{least} $d$-closed set,
which is what the $+d$ inference rule computes.
For example, if $D$  is $p \Rightarrow p$ then $\{p\}$ is a $\partial$-closed set,
but the least $\partial$-closed set is $\emptyset$.
Thus an over-approximation is less directly useful than an under-approximation.
Nevertheless, it can be useful in answering queries in the negative:
if $q$ is not in some over-approximation, then we can be sure that $+d \: q$ is not inferred by $d$.

We have only a single result on over-approximation by $\pl$.

\begin{proposition}     \label{prop:contain6}
Let $D$ be a fact-deficient finite propositional defeasible theory
with empty superiority relation.  Then
\begin{itemize}
\item
$+\delta \phantom{^*} \subseteq +\pl$
\item
$+\delta^* \subseteq +\pl^*$.
\end{itemize}
\end{proposition}
\skipit{
\begin{proof}
Let $d \in \{\delta, \delta^* \}$.
The proof is by induction on the length of proofs in $DL(d)$.
By Corollary~\ref{cor:supplambda}, $\supp_{d} = \lambda$ for $D$.
Now we can apply the same proof approach as Alternative Theorem~\ref{thm:contain1.2}.
Note that if $-\Delta \non q \in P_\Delta$ then  $+\Delta \non q \notin P_\Delta$ (for 2.2),
and
 if $-\supp_d \alpha \in P$ then  $+\lambda \alpha \notin P_\lambda$ (for 2.3.1),
 using $\supp_{d} = \lambda$.
\end{proof}
}

Unfortunately, there do not appear to be useful conditions that support a similar
over-approximation result for $\partial$ and $\partial^*$.
This might be related to the different treatments of ambiguity in $\partial$ and $\pl$.
Ambiguity blocking inference rules have greater inference strength that the corresponding
ambiguity propagating inference rules.
This makes under-approximation by $\pl$ easier, and over-approximation harder.

\section{Conclusion}

Future applications of defeasible reasoning, such as support for explanation in AI,
 will need to compute with defeasible theories
of a much greater size than current implementations support.
This paper begins an exploration of how a defeasible logic designed for scalability
can be employed to support computational tasks formulated in terms of existing defeasible logics.
There remain many improvements to be made, including the development of:
\begin{itemize}
\item
more precise analyses detecting when conclusions can be approximated by $\pl$, including
better analyses for determining $d$-decisiveness
\item
decompositions of defeasible theories so that components that require unscalable processing
can be isolated and/or executed in an order that limits the effects of the unscalability
\item
transformations that convert a defeasible theory to one that is more amenable to approximation
\end{itemize}
These requirements are similar in style to database query optimization,
even if the technical details are different.
So there is reason to hope that this work can be successful.

There is also the question of extending this approach to other defeasible logics
such as annotated defeasible logic \cite{GM17} and modal defeasible logic \cite{modalrules}.

\textbf{Acknowledgements:}
The author has an adjunct position at Griffith University and an honorary position at UNSW.

\bibliographystyle{splncs04}
\bibliography{sdl3_stripped}

\end{document}